\documentclass{article}

\usepackage[final,nonatbib]{nips_2018}
\usepackage[numbers]{natbib}

\usepackage[utf8]{inputenc} % allow utf-8 input
\usepackage[T1]{fontenc}    % use 8-bit T1 fonts
\usepackage{hyperref}       % hyperlinks
\usepackage{url}            % simple URL typesetting
\usepackage{booktabs}       % professional-quality tables
\usepackage{amsfonts}       % blackboard math symbols
\usepackage{nicefrac}       % compact symbols for 1/2, etc.
\usepackage{microtype}      % microtypography
\usepackage{graphicx}
\usepackage{amsmath}

\usepackage{float}

\usepackage[utf8]{inputenc} % allow utf-8 input
\usepackage[T1]{fontenc}    % use 8-bit T1 fonts
\usepackage{hyperref}       % hyperlinks
\usepackage{url}            % simple URL typesetting
\usepackage{booktabs}       % professional-quality tables
\usepackage{amsfonts}       % blackboard math symbols
\usepackage{nicefrac}       % compact symbols for 1/2, etc.
\usepackage{microtype}      % microtypography

\LetLtxMacro\Oldfootnote\footnote

\title{Scaling Human Activity Recognition to edge devices}

\author{
  Manjot Bilkhu \\
  Department of Electrical and Computer Engineering \\
  University of California San Diego\\
  San Diego, CA 92093 \\
  mbilkhu@ucsd.edu  
   \And
  Hammad A. Ayyubi\\
  Department of Computer Science\\
  University of California San Diego\\
  San Diego, CA 92093 \\
  \texttt{hayyubi@ucsd.edu} 
}

\begin{document}

\maketitle

\begin{abstract}
  Video activity Recognition has recently gained a lot of momentum with the release of massive Kinetics (400 and 600) data. Architectures such as \cite{i3d}\cite{c3d}\cite{ltc}\cite{art-net}\cite{r_21_d} have shown state-of-the-art performances for activity recognition. The one major pitfall with these state-of-the-art networks is that they require a lot of compute. In this paper we explore how we can achieve comparable results to these state-of-the-art networks for devices-on-edge. We primarily explore two architectures - I3D \cite{i3d} and Temporal Segment Network \cite{tsn}. We show that comparable results can be achieved using one tenth the memory usage by changing the testing procedure. We also report our results on Resnet \cite{resnet} architecture as our backbone apart from the original Inception architecture \cite{inception}. Specifically, we achieve 84.54\% top-1 accuracy on UCF-101 dataset using only RGB frames.
\end{abstract}

\section{Introduction}
The domain of Activity Recognition has gained a lot of traction recently - mostly on the account of it's omnipresence in human life and secondly on our ever-increasing computational capability. It is being actively pursued for all sorts of applications like smart homes, human behaviour analysis, sports and even security systems. Owing to the recent success of ConvNets in image classification tasks, one expects video classification to achieve similar performance on large-scale datasets. For action recognition, one needs to pay attention to both spatial cues like pixel intensities, appearance, brightness patterns as well as the temporal dynamics of the scenes across the videos. However, activity recognition in videos face a number of challenges when compared to images.

First and foremost, ConvNets fail to model long-term temporal variations. Most traditional methods focus on using short-term information upto 10 or 15 frames to incorporate temporal dynamics of the scene. Usually ConvNets rely on some recurrent network like Recurrent Neural Networks or Long Short Term Memory Networks to encode temporal information present in the scene. However, these networks are known to require huge amounts of compute power as it is and one cannot think of going beyond 10 or 15 frames with this constraint. Secondly, most existing architectures were designed for trimmed videos. These trimmed sequences have actions that only last for 5 to 10 seconds, which is not the case typically. Hence, these become unrealistic when dealing with real-world scenarios where an action may take place only for a small duration in a video, or, two or more actions performed together might imply a third action. 

It is thus important to come up with a sparse sampling technique that can aggregate information present in different parts of videos. It is thus important to come up with a robust framework which can overcome these challenges and can be deployed on-the-edge. We present in this work two architectures we explored for using on-the-edge - I3D and Temporal Segment Network. 

Specifically, our contributions are:

\begin{itemize}
    \item We show Temporal Segment Networks can work for smaller base architectures well too.
    \item We demonstrate how we can reduce memory usage during test time while maintaining comparable accuracy.
\end{itemize}

\section{Related Work}

We briefly present an overview of representative state-of-the-art architecture for video activity recognition.

\subsection{ConvNet + LSTM}
Conventional activity recognition approaches have relied on using ConvNet's as feature extractors for the spatial domain. They are then followed by models which can learn the dynamics of sequences, like Bi-directional Reccurent Neural Networks or more recently, Long short term memory networks \cite{convlstm}. Architectures like ResNet, VGG-16, Inception-v3 have been among the few architectures which have achieved remarkable results on image classification datasets like ImageNet. Such architectures have also been tried out on videos. As expected, these architectures treat each frame of the video independently and we need to pass the encoded features through a LSTM network to learn the complete spatio-temporal representation of our data.

\begin{figure}[H]
\includegraphics[width=\textwidth, height=175pt]{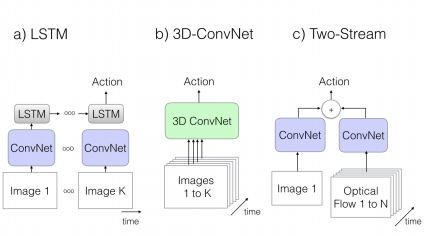}
\caption{Activity Recognition Architectures}
\end{figure}

\subsection{3D ConvNets}
3D ConvNet's appear to be a very natural approach to encode spatio-temporal features because of their hierarchical nature. Their main advantage compared to other approaches is that these networks learn these representations directly. One major drawback of these networks is the considerable increase in the number of parameters, when compared with 2D ConvNet's. Due to this third dimension, it gets pretty hard to train these networks as well as to deploy them on an embedded platform. We have to rely on availability of large datasets to counter these increase in the number of parameters, but as it turns out, even with datasets like Kinetics-400 and UCF-101, we need a lot more data for 3D ConvNets.

\subsection{Two-stream networks}
While LSTM's or RNN's may be able to encode high-level temporal information, they fail to attend to minor changes in sequences. Two-stream networks \cite{twoStreamConv} achieve good response to small motion as well by using two streams: one for the RGB input and the other namely the flow stream. The flow stream is computed by calculating the optical flow of our original video. Optical flow can really capture minute variations in our input and can help us learn low-level motion which can be critical for a lot of cases.

These models have two identical ConvNet's operating on the RGB and flow streams. Both of these ConvNet's are pretrained on ImageNet and provide a good initialization for the two-stream configuration. The RGB and flow stream networks can then be learned in an end-to-end manner independently.  The inputs to the network are 5 consecutive RGB frames sampled 10 frames apart, as well as the corresponding optical flow snippets. This architecture has shown to get very good results on different benchmarks and has been trained end-to-end as well.

\section{Two-stream Inflated 3D Convolutional Networks (I3D)}
Two steam I3D architecture \cite{i3d} builds on the concepts introduced in the architectures above. I3D builds very deep, naturally spatio-temporal feature extractors upon extending state-of-the-art image classification architectures by extending their filters and pooling kernels to 3D. It uses two parallel networks, one each for the flow steam and the RGB stream. The biggest question answered by I3D networks was how to extend weights learned for 2D kernels to 3D. Although I3D networks can achieve great results using just the RGB stream, they also benefit from the two-stream configuration by incorporating the optical flow. 

\begin{figure}[H]
    \includegraphics[width=\textwidth, height=125pt]{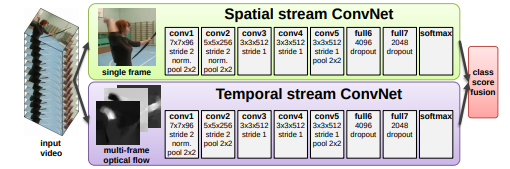}
    \caption{Two-stream I3D Networks}
\end{figure}

\textbf{Inflating 3D ConvNets from 2D ConvNets}. I3D leverages the learned weights from successful ImageNet architectures like Inception-v3, Inception-v1, VGG-16 and ResNet. However, all of these architectures only have 2D kernels. Hence, we inflate each 2D kernel of size N x N to create 3D kernels of size N x N x N. In our experiments, we use Inception-v1 as the backbone architecture from ImageNet.

\textbf{Inflating 3D kernel weights from 2D kernels}. Given that we "inflate" the third dimension for our kernel, the question now arises that how do we initialize the weights across the time axis. Surprisingly, the simplest of approaches works as well, which being, repeat the weights of the N x N 2D kernel N times, and then normalize by dividing with N. This way, the response of our convolutional kernel still remains the same and our model can be implicitly fine-tuned from the ImageNet weights. 

\begin{figure}[H]
\includegraphics[width=\textwidth, height=175pt]{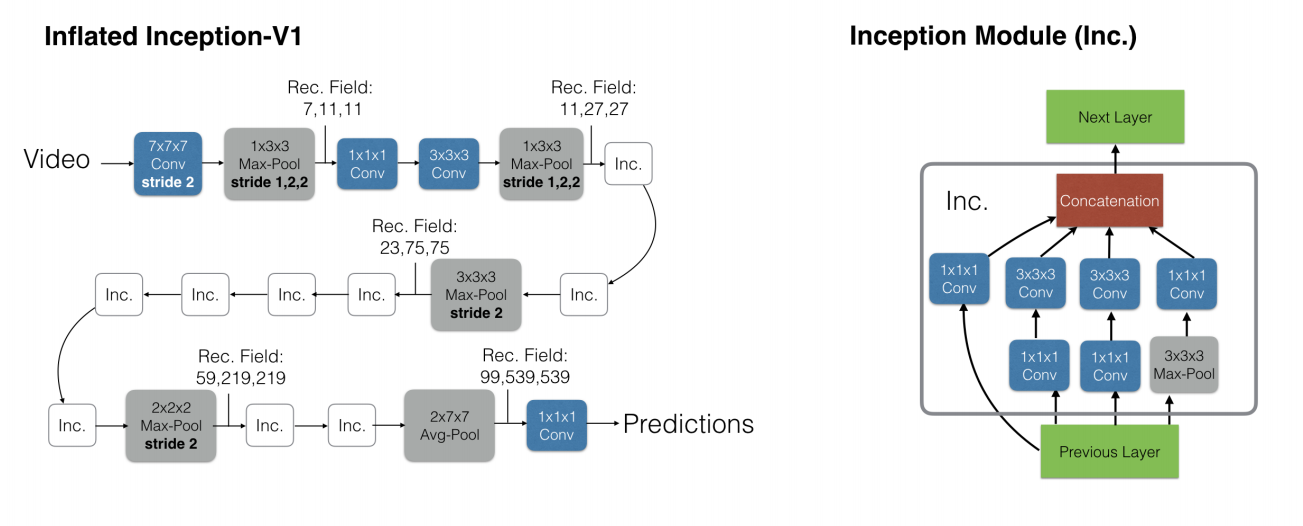}
\caption{Inflating 3D kernels from 2D kernels}
\end{figure}

\section{Temporal Segment Networks}

Temporal Segment Networks \cite{i3d} address one of the primary issues in video activity recognition - modeling long term temporal dependencies in videos. Prior work using ConvNet architectures \cite{twoStreamConv} failed to take this into account or if they did \cite{lstmAction} it was computationally intractable for longer videos. Temporal Segment Network propose a model which addresses these issues by presenting a video level framework.

Precisely, TSN divide the video into $ K $ segments $ \{S_1, S_2, ..., S_K\}$. From each segment a short snippet is randomly sampled. Each snippet makes it's own preliminary prediction about the action in the video. A consensus is then achieved through some consensus function to then combine the scores from these individual predictions. 

\begin{figure}[H]
    \centering
    \includegraphics[width=\textwidth, height=175pt]{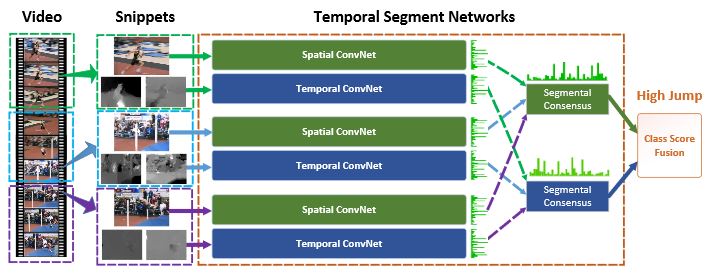}
    \caption{Temporal Segment Network}
\end{figure}

Mathematically, this can be written as:
\begin{equation}
TSN(T_1, T_2, ..., T_K) = \mathcal{H}(\mathcal{G}(\mathcal{F}(T_1;W), \mathcal{F}(T_2;W), ..., \mathcal{F}(T_K;W)))
\end{equation}

$T_1, T_2, ..., T_K$ are the $T_i$ snippet sampled from the corresponding $S_i$ segment. $\mathcal{F}(T_i;W)$ represents a ConvNet operation on $T_i$ with parameters $W$. Important thing to note is that all the snippets share the same parameters $W$. $\mathcal{G}$ is the consensus function used to aggregate predictions from each snippet. $\mathcal{H}$ is the commonly used Softmax function to make predictions.

The negative log logarithmic loss on the final aggregate $G = \mathcal{G}(\mathcal{F}(T_1;W), \mathcal{F}(T_2;W), ..., \mathcal{F}(T_K;W))$,

\begin{equation}
\mathcal{L}(y, G) = - \displaystyle\sum_{i=1}^{C} y_i \left(G_i - log\displaystyle\sum_{j=1}^{j=C}exp(G_j) \right)
\end{equation}

Here, $C$ is the number of classes and $K$ is set to be 3. The aggregate function used is the average function. Although, other aggregate functions like maximum or weighted averaging could also be used. 

Standard stochastic gradient descent (SGD) for backpropagation. The gradient accumulated is of the form:

\begin{equation}
\frac{\partial{\mathcal{L}}}{\partial W} = \frac{\partial \mathcal{L}}{\partial \mathcal{G}} \displaystyle\sum_{k=1}^{K} \frac{\partial G}{\partial \mathcal{F}(T_k)} \frac{\partial \mathcal{F}(T_k)}{\partial W}
\end{equation}

Crucial information in the Eq. 3 is that the gradient updates take into account the information parsed through all the segments. It aggregates the gradients from all the segments. This helps the network in learning the whole video level dependencies rather than a short-sighted frame level dependencies.

\section{Experiments}

As mentioned before, we explored two networks - I3D and Temporal Segment Networks. We list down the results of both sequentially.

\subsection{I3D}

We trained the two-stream inflated 3D convolutional networks with Inception-v1 as the backbone for our experiments. We tested the accuracies obtained on UCF-101 and MSR activity recognition dataset \cite{MSR}. I3D networks were initially introduced along with the kinetics dataset but due to limited compute power and many other restrictions, we were not able to train them end-to-end on the kinetics dataset. UCF-101 dataset has 101 different action categories and 13320 videos in total. The MSR DailyActivity has 16 different classes with 20 videos in each class. 

We first sample the original videos at 25 frames per second and normalize the pixel intensities between 0 and 1. To prepare the flow stream, we use the Lucas-Kanade optical flow method, whose implementation is provided by OpenCV. From these sampled frames, the flow stream is generated and is also normalized between 0 and 1. Note that flow stream only has two components, namely the vector velocities in the x and y direction. For both these datasets, the I3D network was trained with Adam optimizer with a learning rate of 0.002 and a batch size of 16. For the MSR results, due to the limited number of training videos in the dataset, we used pre-trained weights of I3D which are trained on the kinetics-400 dataset and add an additional layer with 16 outputs corresponding to class scores for each of our MSR class. The results are summarized in Table \ref{i3d_results}.

\begin{table}
  \caption{I3D Results}
  \centering
  \begin{tabular}{lll}
    \toprule
    \cmidrule(r){1-2}
    Dataset     & RGB-only     & RGB + Flow ($\mu$m) \\
    \midrule
    UCF-101 & 78.21\%  & 83.76\%     \\
    MSR DailyActivity     & 56.00\% & 57.76\%      \\
    \bottomrule
  \end{tabular}
  \label{i3d_results}
\end{table}

\subsection{Temporal Segment Network}

We perform all our experiments for Temporal Segment Network on UCF-101 dataset \cite{ucf}. We explore various aspects of training categorized as below.

\begin{table}
  \caption{TSN Results}
  \centering
  \begin{tabular}{ll}
    \toprule
    \cmidrule(r){1-2}
    Architecture     & top-1 (RGB only)\\
    \midrule
    Two Stream ConvNet & 72.7\%       \\
    TSN (from scratch)     & 48.7\%       \\
    TSN (ImageNet pretrained)   & 84.5\% \\
    BN Inception TSN (ours) &   84.48\% \\
    ResNet TSN (ours)   &   80.28\% \\
    \bottomrule
  \end{tabular}
  \label{tsn_results}
\end{table}

\subsubsection{Input Modality}

We note that most state-of-the-art architectures for video activity recognition utilize a dual modality for network training - spatial represented by RGB frames and temporal represented by flow fields. Architectures such as \cite{twoStreamConv} \cite{i3d} \cite{tsn} gain significant improvement to their performance by adding flow fields to the RGB frames and then concatenating/aggregating the predictions from RGB frames and flow field in from or the other.

In our case, we narrowed our experiments to RGB frames as our only input modality. This was done to reduce memory usage and time, which is essential for edge-devices. We show in Table \ref{tsn_results}  that good results can still be acheived with only RGB frames.

\subsubsection{Network Architecture}

Original Temporal Segment Network uses Inception-v2 \cite{inception} as their base model. We report our results obtained by training Inception-v2 as our base model as well as Resnet \cite{resnet} as the base architecture. We find that a performance of 81.77\% can still be achieved using a much smaller network architecture such as Resnet.

\subsubsection{Memory usage during Testing}

For the purpose of testing \citet{tsn} divide the video into 25 segments and sample a snippet (frame) from each segment. 10 random crops are then obtained from each frame. Snippet level prediction is made by averaging the prediction from each crop. Next, predictions from each of the 25 frames is combined to give video level prediction. 

In our work, we only divide the video into 25 segments, sample one frame from each of the segment and make frame level from only one crop. Now the frame level predictions from each of the 25 segments is aggregated to give the video level prediction. This small but subtle difference allows us to reduce the memory usage 10 times as compared to the work by \citet{tsn}. This becomes very crucial for edge devices. We report this in table \ref{tsn_memory}.

\begin{table}
  \caption{TSN Memory Usage}
  \centering
  \begin{tabular}{lllll}
    \toprule
    \cmidrule(r){1-2}
    Architecture  & Batch Size  & \# Crops    & top-1 (RGB only) & Memory (G)\\
    \midrule
    TSN  & 128  & 10  & 84.5\% & 17\\
    BN Inception TSN (ours) & 128   & 1 & 84.48\% & 1.77 \\
    BN Inception TSN (ours) & 32    & 1 & 84.54\% & 1.8 \\
    \bottomrule
  \end{tabular}
  \label{tsn_memory}
\end{table}

\section{Conclusion}

In this work we explored two popular architectures for video activity recognition - I3D and Temporal Segment Networks. We demonstrated that Temporal Segment Networks can work for smaller base models such as ResNet as well. This result hints that we may be able to reach perhaps better results with some external guidance such as "Dark Knowledge" \cite{kd}. Secondly, we may be able to reduce the model size even further while maintaining the accuracy level. 

We were successful in reducing the memory usage $\sim$ 10 times by making subtle changes during test time. We brought the memory requirement to within 2G for temporal segment networks to work. This result is very crucial for the purpose of running it on edge devices.

\bibliographystyle{plainnat}

\bibliography{nips_2018}

\end{document}